\documentclass[11pt]{article}
\usepackage{coling2016}
\usepackage{times}
\usepackage{url}
\usepackage{latexsym}
\usepackage{graphicx}
\usepackage{enumitem}
\setitemize{noitemsep,topsep=0pt,parsep=0pt,partopsep=0pt}
\newcommand{\kcr}{{\em KCR}}
\newcommand{\rcr}{{\em RCR}}
\newcommand{\bcr}{{\em BCR}}
\newcommand{\dcr}{{\em DCR}}

\newcommand{\kcrb}{{\em KCR }}
\newcommand{\rcrb}{{\em RCR }}
\newcommand{\bcrb}{{\em BCR }}
\newcommand{\dcrb}{{\em DCR }}

\title{Veracity Computing from Lexical Cues and Perceived Certainty Trends}

 \author{
 Uwe D. Reichel\thanks{$^*$UDR is supported by an Alexander von Humboldt Society grant.}\\
       Research Institute for Linguistics\\
        Hungarian Academy of Sciences\\
        Budapest, Hungary\\
        {\tt uwe.reichel@nytud.mta.hu}
         \And
 Piroska Lendvai\thanks{$^\dagger$PL is supported by the PHEME FP7 project (Grant No. 611233).}\\
        Computational Linguistics\\
        Saarland University\\
        Saarbr\"ucken, Germany\\
        {\tt piroska.r@gmail.com}\\
  }
\date{}

\begin{document}
\maketitle
\begin{abstract}
We present a data-driven method for determining the veracity of a set of rumorous claims on social media data. 
Tweets from different sources pertaining to a rumor are processed on three levels: first, factuality values are assigned to each tweet based on four textual cue categories relevant for our journalism use case; these amalgamate speaker support in terms of polarity and commitment in terms of certainty and speculation. Next, the proportions of these lexical cues are utilized as predictors for tweet certainty in a generalized linear regression model. Subsequently, lexical cue proportions, predicted certainty, as well as their time course characteristics are used to compute veracity for each rumor in terms of the identity of the rumor-resolving tweet and its binary resolution value judgment. The system operates without access to extralinguistic resources. Evaluated on the data portion for which hand-labeled examples were available, it achieves .74 F1-score on identifying rumor resolving tweets  and .76 F1-score on predicting if a rumor is resolved as true or false.
\end{abstract}

\section{Background and Task Definition}
\label{sec:intro}

A growing amount of studies investigate how rumors and memes spread and change on social media platforms \cite{Leskovec,Qazvinian2011,Procter2013}; given the amount of user-generated content,  the need for automatic fact checking and claim verification procedures is obvious. 
To compute veracity, systems have been created recently for assessing the credibility of sources and claims \cite{berti_veracity}.  
Upcoming initiatives endorsed  veracity detection in social media content as a shared task, calling for  targeted applications and  releasing benchmark data\footnote{http://alt.qcri.org/semeval2017/task8/}.

To tackle this challenge, we implemented a system that seeks to achieve three goals: (i) to compute a judgment indicating how factual a claim is, based on textual cues and predicted speaker certainty, (ii)  to identify which tweet  is resolving a rumor, in a set of tweets that discuss this rumor, and (iii) to predict the resolution value for the rumor, i.e., whether the rumor is verified as true or false.
Veracity computation is based on information from three information layers related to rumorousness: (1) lexical-level factuality  cues, (2) temporal patterns, and  (3) speaker certainty.  
The system is purely data-driven and operates without building claim source profiles for the analyzed content.
Below we introduce our motivation in the context of previous and related work.

The means by which factuality is conveyed are largely but not exclusively encoded on  linguistic levels  and are tightly related to the notion of certainty. 
Certainty and other  extra-propositional aspects of meaning
have prominently been investigated in terms of modality, negation and speculative language phenomena \cite{morante_scope,exprom}. 
Benchmark corpora with annotations emerged  \cite{factbank,Farkas_sharedtask}, and systems have been built \cite{sauri2012,deMarneffe_didithappen,Velldal} to process texts from the genres of literature, newswire, biomedicine and online encyclopedia, typically drawing on  lexical and syntactic cues. 
\cite{cross_uncertainty} propose a  method for porting uncertainty detection across genres and domains.
\cite{kilicoglu} present a full-fledged, compositional approach to factuality modeling and detection on   texts from the domain of biomedicine based on fine-grained typology and dictionary-based classification of extra-propositional phenomena.
Several components of the model are motivated by the nature of scientific communication that serves to
track hypothesis building processes with tentative results,
analogously to 
 journalistic reports about breaking news.
\cite{Soni14modelingfactuality} focus on factuality framing in social media data in quoted claims with a small set of cues, whereas \cite{twittertrails}
implement keyword-based negation detection without providing quantitative evaluation. 

Next to linguistically expressed uncertainty,
extralinguistic information such as the temporal distribution of claims is shown to be an important aspect of veracity computation.
Previous studies that investigated  temporal patterns of linguistic cues tied to claims emerging in real-world events focus on keywords related to  sentiment, named entities and domain terms \cite{temnikova_avarga}, but not factuality-conveying cues.
\cite{wei_sm_uncert} report on the first uncertainty corpus based on tweets, as well as on classification results for uncertain tweets. Next to platform-specific metadata, they utilized 
cue phrases  in annotated uncertain tweets and an 
algorithm to 
detect peaks in the data.
\cite{Kwon2014} and \cite{Ma2015} show for rumor detection that 
accuracy can be improved by not only looking at message-related properties but also at how these properties change over time.  \cite{Ma2015} propose a time series structure for features and their deltas as the input for classification. 

On the full PHEME dataset, \cite{Lukasik} report on stance detection in the context of  temporal dynamics. 
They utilize textual information via language modeling but do not evaluate the contribution of textual as opposed to other features.
On the same dataset, \cite{Zubiaga_plosone} analyzed labeled certainty values in dependence of claim resolution, and
found that
tweeters post messages with statistically similar  certainty before and after a claim is resolved, moreover, irrespective of the resolution value.

In \cite{success16} we analyzed and validated a subset of the PHEME data on English and German data that temporal distribution and polarity of lexical markers can be used to represent and quantify changes in factuality framing in a rumor's lifecycle. 
Our current study  furthers this research by 
incorporating, evaluating, and visualizing temporally anchored 
features for  claim resolution point as well as claim resolution value prediction in English language rumors discussed in potentially noisy, user-generated content.

The paper is structured as follows. 
In Section \ref{sec:data} we
introduce the underlying  data and certainty annotations,
and describe the automatic extension of lexical cues 
assigned to four levels of factuality.
In Section \ref{sec:facPred} the relation between certainty and each of the factuality levels is assessed, 
and  regression analysis is used for
predicting certainty values by cue-type ratios. In Section \ref{sec:facDriftQuant} 
we quantify trend discontinuities in time series data of lexical cue ratios and predicted certainty scores to describe rumor resolution points.
Cue ratios, certainty, as well as their time course characteristics
are exploited  in
Section \ref{sec:facDriftPred}, where we train classifiers to identify
claim-resolving tweets within series of tweets spanning a claim's lifetime, and additionally predict the claim's resolution value. 
The findings are discussed in Section \ref{sec:conc}.

\section{Data }
\label{sec:data}
\subsection{Corpus}
\label{sec:crp}

We worked on a subset of a freely available, annotated social media corpus\footnote{https://figshare.com/articles/PHEME\_rumour\_scheme\_dataset\_journalism\_use\_ case/2068650} collected from the Twitter platform\footnote{twitter.com}, containing  tweets in English related to 
 three crisis events: the Ottawa shooting\footnote{https://en.wikipedia.org/wiki/2014\_shootings\_at\_Parliament\_Hill,\_Ottawa}, the Sydney Siege\footnote{https://en.wikipedia.org/wiki/2014\_Sydney\_hostage\_crisis}, and the Germanwings crash\footnote{https://en.wikipedia.org/wiki/Germanwings\_Flight\_9525}.  
Each event is annotated in terms of several rumorous claims\footnote{We use {\it rumor}, {\it rumorous claim}, and {\it claim} interchangeably to refer to the same concept.} -- plausible but at the time of emergence unconfirmed statements, e.g.\ in the Sydney Siege collection two example claims are
 "{\sf  \small There is a hostage situation at a cafe in Sydney}" and
 "{\sf  \small Police (authorities) have been in contact with the hostage-taker}".
For each claim, there are a set of tweets that discuss or mention that claim, and a single one of these tweet has been manually identified and judged to be 
authoritatively resolving the claim either as true or false.
A resolving tweet  for the claim  "{\sf  \small The Germanwings plane experienced a rapid descent before crashing}"
is: {\sf  \it  \#4U9525 took eight minutes to descend from 38,000 feet to impact, says Germanwings CEO Winkelmann};  this rumor is annotated by a journalist as resolved True.
The verification value is inherited by all the tweets that relate to this claim, also in retrospect.
To ensure that there is always exactly a single resolving tweet per claim we discarded the unresolved claims in the given corpus so that the data underlying this study amounts 45 claims containing in sum 11,420 tweets.
Tweets are organized into threaded conversations and are marked up with respect to 
seven categories of evidence, among others stance and certainty;
for full details on the corpus we refer to \cite{Zubiaga}.

\subsection{Certainty annotations}
\label{subsec:facAnnot}

Certainty annotations were pre-assigned in the corpus in relation to stance value annotations by \cite{Zubiaga}. 
Stance represents speaker attitude toward a target: in this corpus, the target is a rumorous claim, and
 each tweet was manually marked as either supporting, denying, questioning, or commenting a claim.
 Tweets that received either of the stance labels {\it supporting} or  {\it denying} were additionally assigned a certainty value. This value served to express  tweeter confidence with respect to their stance, as perceived by independent, crowdsourced annotators. Each tweet  was annotated by 5-7 crowdsourced annotators,  in terms of the four labels {\em uncertain, somewhat certain, certain}, and {\em underspecified}.  

We further processed these annotations as follows. To cope with frequent annotation mismatches, we did not simply pick the majority-voted certainty label for our subsequent analyses, but aggregated the annotated values for each tweet as follows. 
The original certainty labels were mapped to the numerical values 0, 1, 2, and {\em NaN},
respectively. 
We then calculated the mean of all non-{\em NaN} values
and normalized this to the interval between 0 and 1 by dividing it by
the maximum score 2.
For example, the tweet {\it "Now hearing 148 passengers + crew on board the \#A320 that has crashed in southern French Alps. \#GermanWings flight. @BBCWorld"}  was labeled as 'certain' by 4 annotators and labeled as 'somewhat-certain' by 1  annotator, so this tweet was assigned by us the certainty score of 0.9.
   
The intersection of tweets that relate to claims that were not only annotated as relating to a resolved claim, but also annotated with the three utilizable  certainty labels {\em uncertain, somewhat certain, certain} left us with a relatively small amount of tweets (266),
while we also had at our disposal a larger set of  tweets (946) with claim resolution annotation but no certainty values assigned. We made use of both collections as described below.

\subsection{Factuality cues: from seeds to extended lists}

The material underlying our study is user-generated content. The data collection method, cf. \cite{Zubiaga} retained only microposts that passed a retweet count threshold, often by media outlets using well-formed language. 
Since replying tweets 
are also included in the corpus, a large portion of the data involves noisy texts. 
Based on the factuality literature, most prominently \cite{factbank} and \cite{Soni14modelingfactuality}, we devised four factuality groups, each holding up to 40 single-token lexical cues. There is no restriction on which part of speech category a cue may belong to.

\vspace{.1in}

\begin{itemize}
\item {\bf Knowledge} cues, e.g.  clarify, confirm, definitely, discover, evidence, explain, official \ldots
\item {\bf Report} cues, e.g. according, capture, claim, footage, observe,  report, say, show,  source \ldots
\item {\bf Belief} cues, e.g. apparent, assume, believe, consider, perhaps, potential, presume, suspect \ldots
\item {\bf Doubt} cues, e.g.  ?, accuse, allege,   contrary, deny, incorrect, misstate, not, unsure, why, wrong \ldots
\end{itemize}

\vspace{.1in}
 Each group represents one complex aspect of factuality that can be intuitively understood by non-linguists, i.e. end users  in the journalistic verification scenario. 
Knowledge and belief cues express affirmative factuality polarity on graded levels of certainty.
Report cues express affirmative factuality polarity as well, and additionally  delegate  speaker commitment, as they typically occur in externally-attributed statements and evidence,  indicating a stronger level of speculation than knowledge cues.
Doubt cues express negative factuality polarity and were selected to be indicative of contradictory or opposing-stance statements which can be extremely strong signals for rumorous claims.
 Involving categories that have been suggested in previous work in  fully-fledged factuality taxonomies  (see Section 1) would require extracting higher-level linguistic information such as dependency parses, which are difficult to obtain from noisy data, and are  reserved for the extended version of our system.

Starting from the seed cues, each of the four lists were automatically further populated from available semantic resources: via extracting the top-3 most similar items from the pretrained Google News word embeddings vector\footnote{http://code.google.com/p/word2vec/}, as well as lemmas from the top-3 synsets from the English WordNet \cite{Fellbaum1998} via NLTK\footnote{http://www.nltk.org/\_modules/nltk/corpus/reader/wordnet.html} \cite{Bird2009}. Only single-token items were harvested; each cue token was subsequently extended by its derivationally related forms via the corresponding NLTK function.
Using the extended trigger lists, we obtained counts for each tweet via matching each cue list to a tweet's content, applying the NLTK Snowball Stemmer\footnote{http://www.nltk.org/api/nltk.stem.html} prior to lookup. We have also experimented with syntax-based cue disambiguation, but opted to abandon it for no proven impact on our current task setup.  To exemplify cue matching, in the tweet 
"{\it \#BREAKING: @nswpolice say a photo circulating of arrest of man near \#MartinPlace is NOT related to the police operation \#sydneysiege}" our lookup matches two 'report' cues ({\it say} and {\it photo}) and one 'doubt' cue ({\it not}).
The cue extension procedure boosted the seed lists with a few hundred new tokens per cue group, leading to more cue matches in tweets. 

\begin{figure}[!t]
\begin{center}
\includegraphics[scale=.37]{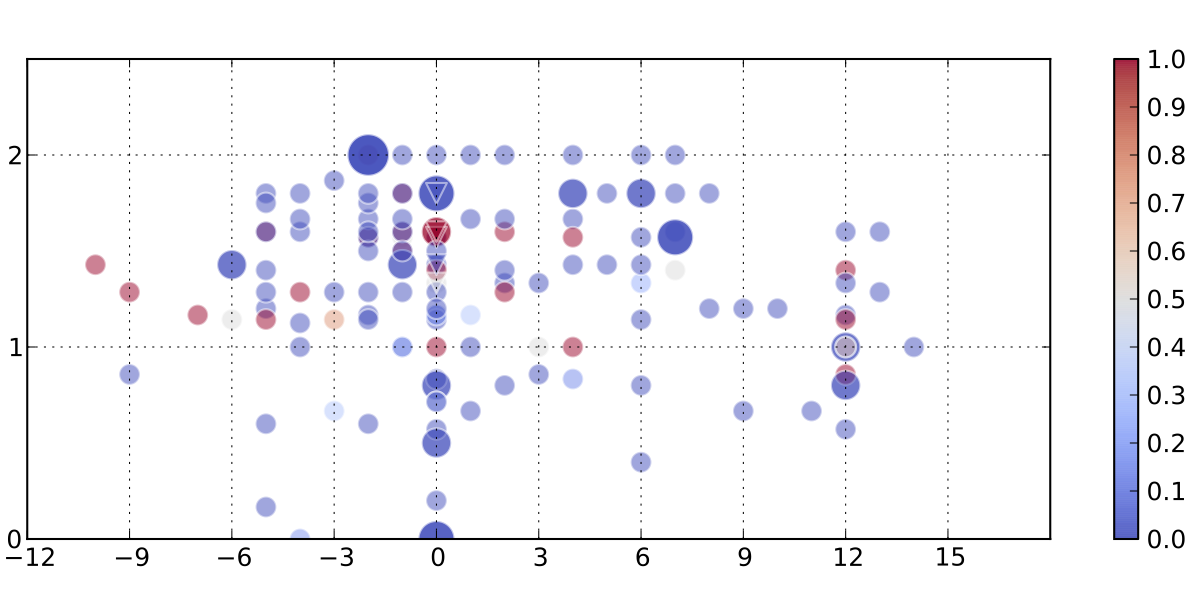}
\hfill
\includegraphics[scale=.37]{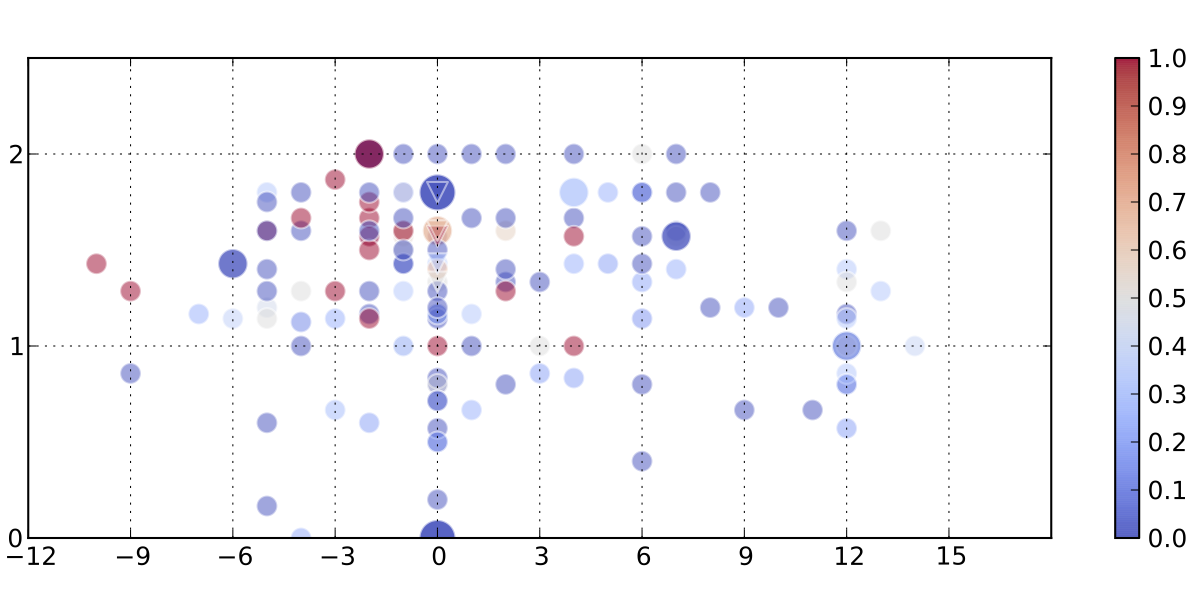}
\end{center}
\caption{{\it Color-coded factuality cue ratio (FCR)} 
derived from  matched cues in the two affirmative-polarity and high-commitment featuring factuality cue groups ('knowledge cue ratio' + 'report cue ratio'),
using seed cues (left) vs extended cues (right). Factuality ratios are plotted on perceived certainty patterns (cf. Section \ref{subsec:facAnnot}) in 163 argumentative Twitter microposts related to 5 rumorous claims resolved as {\it True} during Sydney Siege.  x axis: 10-minute intervals normalized to {\it claim resolution point  at 0} (resolving tweets marked by triangles), y axis:  {\it certainty score} averaged over perceived certainty judgments. Partial time window, size-coded amount of evidence.}
\label{fig:timeline}
\end{figure}

 Arguably, there are cues that might belong to more than one factuality group, most prominently  negation words  that, depending on their scope, may  express  certainty as well.  We  
hypothesized however that utilizing the contextual distribution of a cue will
represent its certainty-encoding function in rumor resolution timelines in a robust way.
We exemplify such a timeline in Figure \ref{fig:timeline}, differentiated by certainty values plotted to the y axis. The increase in cue recall based on extended lists aimed to benefit the certainty and rumor resolution modeling steps that we introduce in the next sections.
 
 \paragraph{Features derived from matched cues}

Based on the extended cue matching counts,  
for each tweet we calculated the proportion of cues for each factuality  group over all cues: 
each proportion ranges from
0 (no cue of the respective type) to 1 (cues exclusively of the respective type). Each tweet is thus represented by the four ratios \kcrb  ('knowledge' cue ratio), \rcrb  ('report' cue ratio), \bcrb 
('belief' cue ratio),  \dcrb  ('doubt' cue ratio). 
The   \rcrb  for the above example tweet is 2/3, the \dcrb 1/3, while  \kcrb and  \bcrb are 0.

\section{Certainty prediction}
\label{sec:facPred}
 
Next, we examined the relation of cue ratios to certainty values assigned to tweets via regression analysis.

\paragraph{Method}

To predict the degree of certainty for each tweet, we fitted
generalized linear models (GLM) to the tweets manually annotated for perceived certainty.
The four
lexical cue ratios were defined as predictors and the certainty score (see Section \ref{subsec:facAnnot}) as  target. To restrict the output to interval $[0$ $1]$, the distribution of the response was set to binomial, and a logit link function was chosen. 
A zero inflation problem is given due to 
the frequent absence of cue words,
which we addressed by adding observation weights to the data points as follows: for each of a predictor's values the variance of the associated target values was measured, normalized by the variance sum, and  its inverse taken.
Zero values of a predictor co-occurring with a high variance of certainty values thus received a low weight when fed into the regression. The weight of each feature vector was then derived by taking the mean of the predictor-related weights.

\paragraph{Results}

Spearman's Rho correlations between the single lexical cue ratios and the normalized certainty score are small. Only for \kcrb ($.14$) and \dcrb ($-.39$) the correlations are significant (Wilcoxon two sided signed rank tests for paired samples, $p<0.05$). These two correlations point in the expected direction.
The 10-fold cross validation of the GLMs on held out data yielded an
average root mean square error of 0.22 (maximum: 0.28), which is significantly lower than the error of the baseline model always predicting the observed mean
certainty value (two-sided Wilcoxon signed rank test for paired
samples, $p<0.01$). Finally, we fitted a GLM to all available training data for certainty prediction of all tweets used for the subsequent claim resolution prediction and resolving tweet localization tasks.

\section{Certainty trend quantification}
\label{sec:facDriftQuant}

\paragraph{Method}

In our approach we  address time course characteristics more explicitly than previous studies \cite{Kwon2014,Ma2015}. That is, instead of bundling feature vectors at different time stamps to a joint vector, we capture time course characteristics by parameters of regression lines fitted through the feature values over time. These regression lines will be used for trend discontinuity analyses as described in detail below. By means of this analysis we augmented the set of
variables the following way: for all four cue ratios and the derived
certainty score we calculated four discontinuity features yielding
five features for each cue ratio and certainty. These variables will
be introduced in the following paragraphs and represent:

\vspace{.05in}

\begin{itemize}
\item tweet-intrinsic properties (lexical cue ratios \kcr, \rcr, \bcr, \dcr, predicted certainty {\em CRT})
\item local discontinuities across pairs of subsequent tweets ({\em
  *\_Delta})
\item global discontinuities in linear cue and certainty trends ({\em *\_Reset, *\_RMSD\_p, *\_RMSD\_f}), where the asterisk stands for the tweet-intrinsic variables, i.e. the cue ratios and certainty.
\end{itemize}

\vspace{.05in}

\begin{figure}[t]
\begin{center}
\includegraphics[height=4cm, width=0.8\textwidth]{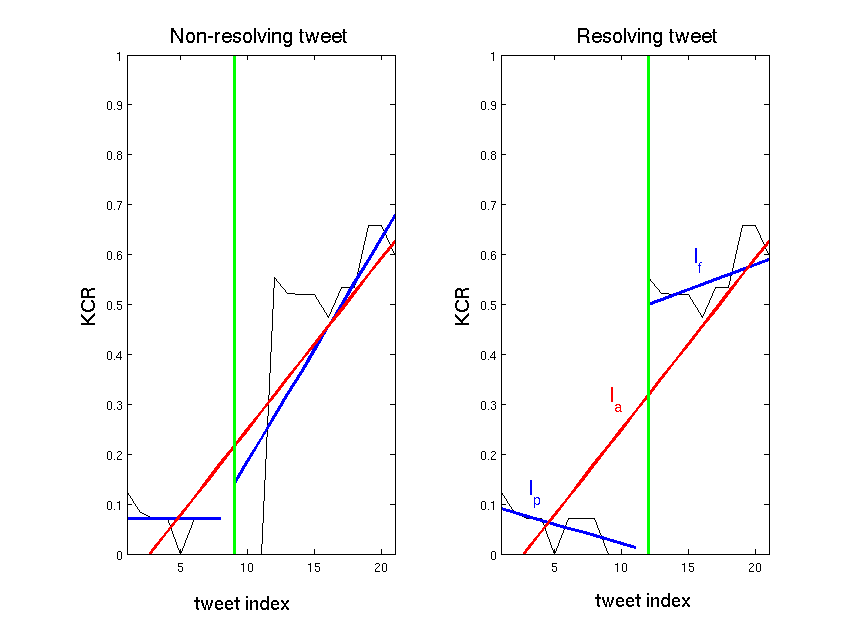}
\end{center}
\caption{KCR trend analysis. For each tweet (green vertical line) three regression lines are fitted to the \kcrb sequence (black): to the preceding and the following sequence ($l_p$, $l_f$; blue), and to the entire rumor ($l_a$; red), the latter representing the general trend. For resolving tweets (right) $l_p$ and $l_f$ deviate more from a common trend, which is expressed by a larger reset as well as by larger root mean squared deviation values from the overall trend line $l_a$.}
\label{fig:styl}
\end{figure}

For measuring discontinuities, tweets were indexed in the order of their time stamps. Local discontinuities are measured in terms of the delta deviations of each tweet $i$ from the
preceding tweet, i.e. by subtracting each intrinsic variable's value
of tweet $i-1$ from the corresponding value of tweet $i$ ({\em
  *\_Delta}).  To quantify the amount of discontinuity a tweet induces
in the overall trend of a variable, we fitted three regression lines:
line $l_p$ through the intrinsic values of the preceding tweet
sequence $1 \ldots i-1$, line $l_f$ through the values of the
following tweet sequence $i+1 \ldots n$ ($n$ be the number of tweets
in a claim), and line $l_a$ through the entire tweet sequence $1
\ldots n$.  The method is illustrated in Figure \ref{fig:styl}.

In
order to measure the amount of discontinuity for each intrinsic
variable at each tweet, we calculated the reset, i.e. the difference between the offset of $l_p$ and the onset of $l_f$ ({\em
  Reset}), and the deviation of each of these lines $l_p$ and $l_f$
from $l_a$ in terms of root mean squared deviation ({\em RMSD\_p},
and {\em RMSD\_f}, respectively). The method was adopted from
intonation research \cite{RM_IS2014}, where it is used to
quantify pitch discontinuities for prosodic boundary strength
prediction. 

 Applying the reasoning of \cite{RM_IS2014}, {\em
  Reset} quantifies the disruption at each tweet, and {\em
  RMSD\_p,f} quantify the deviation of the tweet preceding and
following regression lines from a common trend.  Figure \ref{fig:styl}
gives an example how the regression lines preceding and following a
tweet deviate less from a common trend for non-resolving tweets (left
half) than for resolving tweets, expressed by lower values for {\em
  Reset, RMSD\_p}, and {\em RMSD\_f}. This example therefore
illustrates a higher impact of the resolving tweet on the
claim-level trends.

\paragraph{Results}

\begin{figure}[b]
\begin{center}
\includegraphics[width=0.83\textwidth]{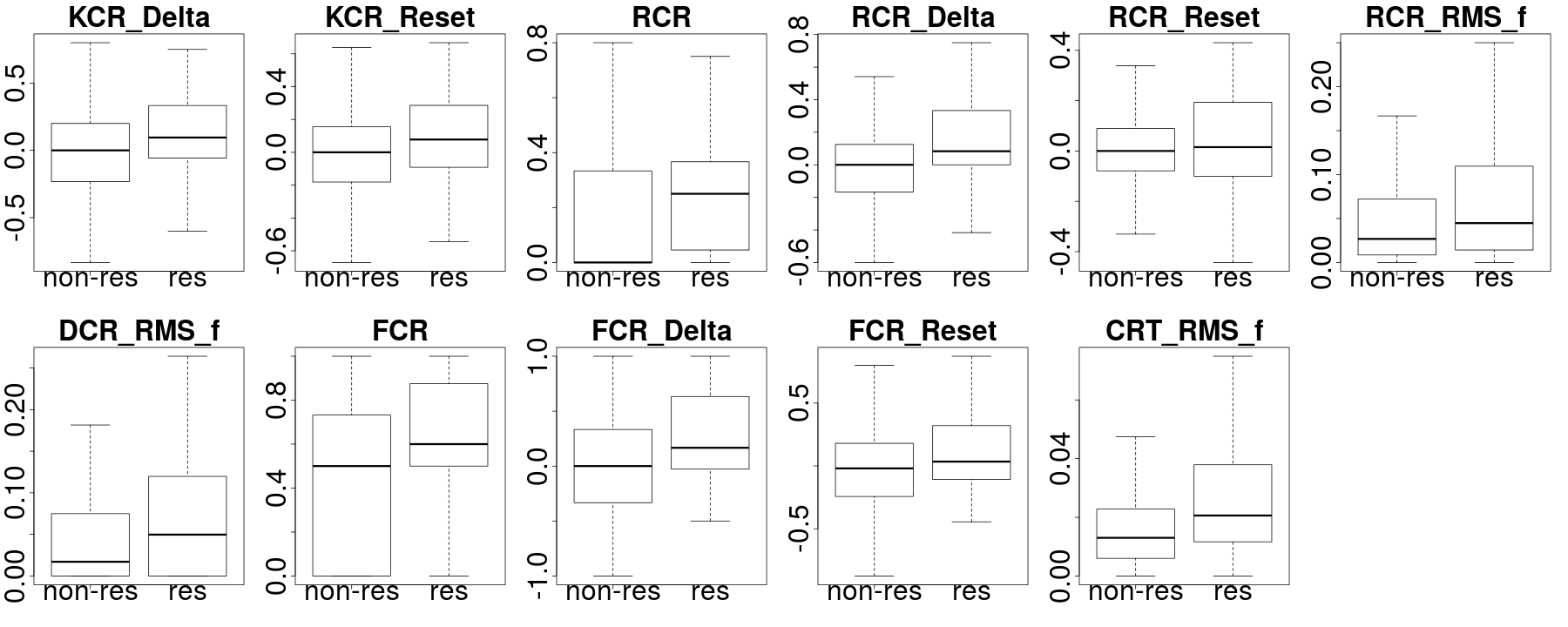}
\end{center}
\caption{Significant differences between non-resolving ({\em non-res}) and resolving ({\em res}) tweets. Resolving tweets show higher values for tweet intrinsic properties (RCR, FCR=KCR+RCR), their local discontinuities (*\_Delta), and their global trend discontinuities (*\_Reset, *\_RMS\_f).}
\label{fig:stat_res}
\end{figure}

For all five tweet-intrinsic measures as well for the related sets of
four derived local and global discontinuity measures we tested the
difference between resolving and non-resolving tweets by linear
mixed-effect models with each of the measures as dependent variable,
$+/-$ {\em resolving tweet} ({\em RES}) and {\em Rumor is Resolved as True vs False} ({\em VAL}) as the fixed effects, and {\em event} as random effect. Due to the large number of tests p-values were corrected for false discovery rate \cite{Benjamini2001}. All significant feature differences (after p-value correction, $p<0.05$) for {\em RES} are shown in Figure \ref{fig:stat_res}.

\begin{figure}[!ht]
\begin{center}
\includegraphics[width=0.7\textwidth]{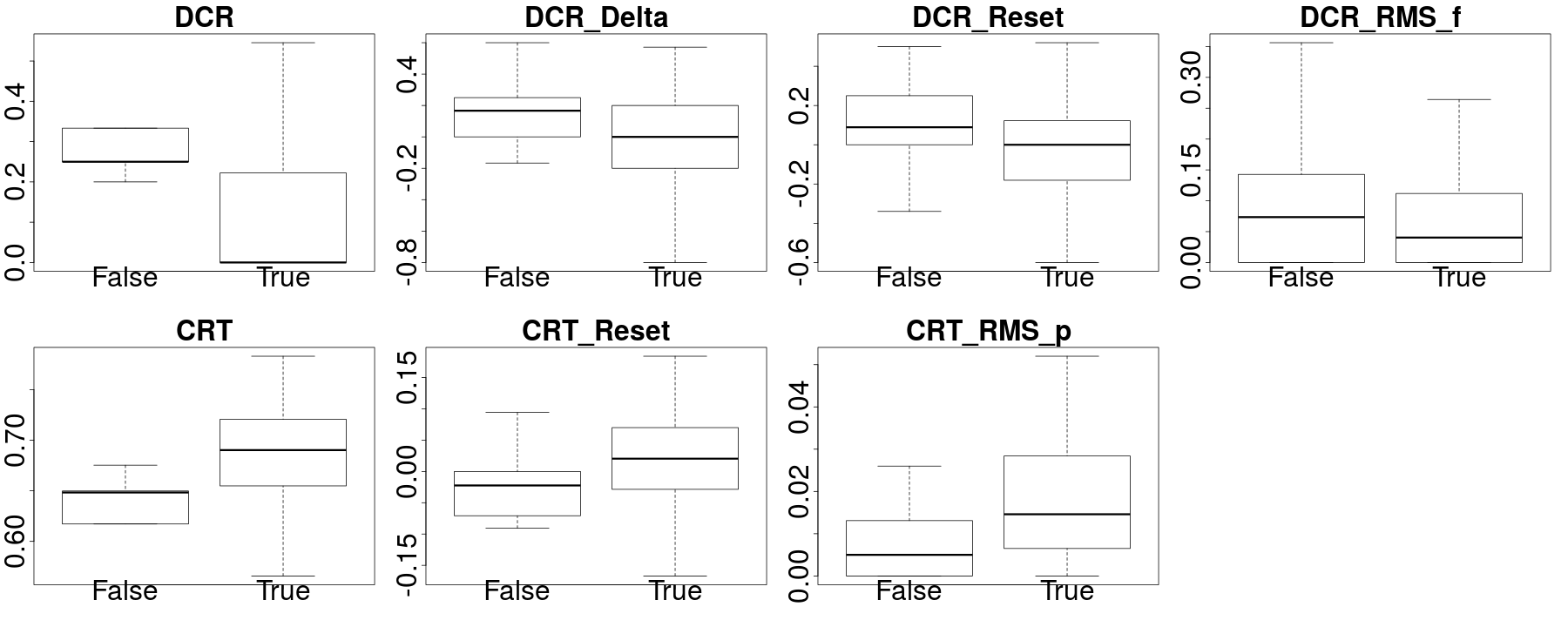}
\end{center}
\caption{Significant differences for resolving tweets in claims validated as {\em True} vs {\em False}. In {\em verified} claims, resolving tweets show higher certainty cue (CRT) related values than non-resolving tweets. In {\em falsified} claims, resolving tweets show higher doubt cue (DCR) related values than non-resolving tweets.}
\label{fig:stat_val}
\end{figure}

\begin{figure}[b]
\begin{center}
\includegraphics[height=2.6cm, width=0.8\textwidth]{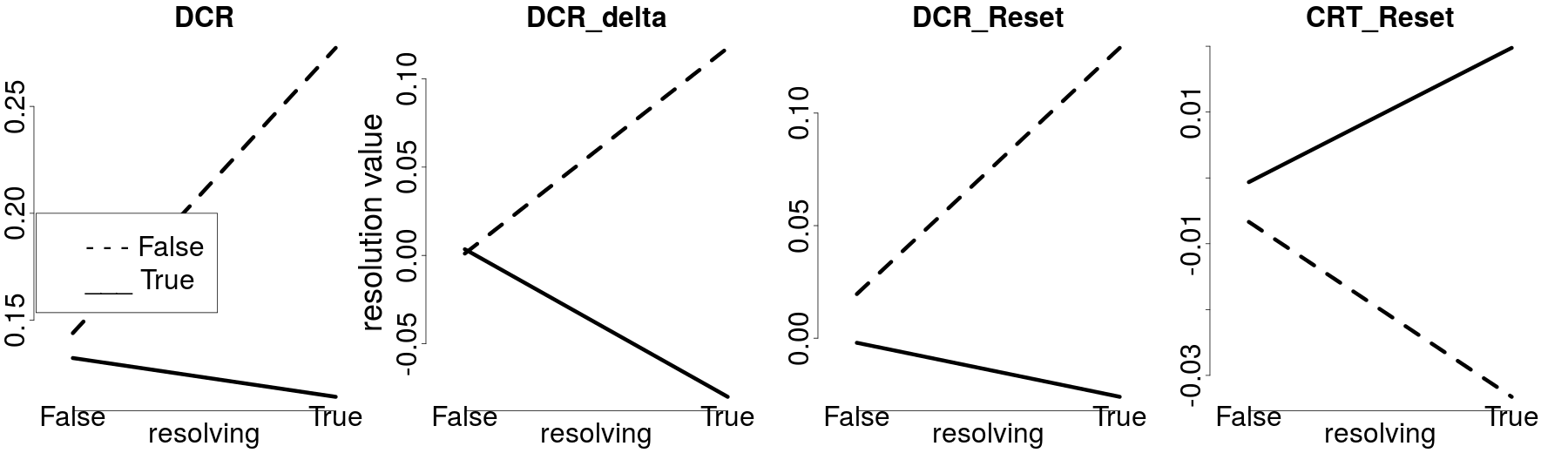}
\end{center}
\caption{Interactions of the effects {\em Resolving} and {\em Resolution value}. For resolving tweets in verified and falsified claims, certainty (CRT) and doubt (DCR) cues behave in an opposite way; see Discussion.}
\label{fig:interact}
\end{figure}

The claim resolution value {\em VAL} turned out to affect the variables related to doubt cue ratio and to predicted certainty which is shown in Figure \ref{fig:stat_val} ($p<0.01$). Significant interactions between {\em RES} and {\em VAL} solely affect the doubt cue and certainty variables and are presented in Figure \ref{fig:interact}. All reported findings are discussed in Section \ref{sec:conc}.

\section{Predicting the resolving tweet and its resolution value}
\label{sec:facDriftPred}

As illustrated in Section \ref{sec:facDriftQuant} the distinction of tweets in rumor resolving and non-resolving as well as their resolution values have an impact on several of the examined cue ratio, certainty and discontinuity variables. Our next step thus was to use these variables to predict:

\begin{itemize}
\item for each tweet whether it resolves a rumor or not ({\em RES}),
\item for each resolving tweet, whether its resolution value is True or False ({\em VAL}).
\end{itemize}

\paragraph{Method}

For both binary classification tasks we enlarged the feature vector for each tweet by {\em tweet density}, i.e. the mean number of tweets per minute in a 10 minutes time window centered on the respective tweet. As for the features introduced above, also together with tweet density its four discontinuity measures (cf. Section \ref{sec:facDriftQuant}) were added. We then subdivided the features into two sets:

\begin{itemize}
\item {\em CueSet:} consisting of all lexical knowledge, report, belief, and doubt cue ratios, their derived discontinuities, as well as tweet density and its discontinuities.
\item {\em CertSet:} consisting of the predicted certainty values (cf. section \ref{sec:facPred}) and their discontinuities instead of the raw cue ratios, and of the tweet density features.
\end{itemize}

This division serves to test whether raw cue ratio or derived certainty features are better suited for the two classification tasks. Task {\em RES} is carried out on all tweets in our data, whereas task {\em VAL} only applies to resolving tweets. Both tasks were carried out and evaluated in isolation to independently assess the respective performance. That is, the training and testing items for {\em VAL} were not taken over from the preceding {\em RES} classification output but from the original data set.

Since in both data sets the target value distributions are highly skewed, we applied resampling without replacement to avoid overlaps of training and test items in subsequent 10-fold cross-validation. The maximum sample size was determined as a weighted mean of the given sample sizes to ensure that for {\em VAL} and for {\em RES} the more
frequent class occurs maximally twice as often as the less frequent one. By this resampling the amounts of claims and tweets (cf. section \ref{sec:crp}) were reduced to 39 (13 falsified, 26 verified claims) and 138 (46 resolving, 92 non-resolving tweets), respectively.

We then applied AdaBoostM1 classifiers \cite{Freund1999} (Matlab function {\em fitensemble}, 40 weak learners, minimum 2 items per leaf and 3 items per non-final node) to both data sets and comparatively evaluated the results in a ten-fold cross validation.

\paragraph{Results}

Table \ref{tab:res_pred} summarizes the mean performance values on the held-out data after cross-validation. {\em BL Accuracy} represents the baseline performance which is defined as predicting only the most frequent class and is quite high due to the not entirely resolved skewedness. The results are discussed in the next section.

\begin{table}[t]
\begin{center}
\begin{tabular}{|l|l|l|l|l|l|l|l|l|}
\hline
Task & Feature set & wgt F1 & wgt Recall & wgt Precision & Accuracy & BL Accuracy \\
\hline
{\em RES} & CueSet & 0.74 & 0.73 & 0.74 & 0.77 & 0.67 \\
{\em RES} & CertSet & 0.69 & 0.71 & 0.75 & 0.70 & 0.67 \\
\hline
{\em VAL} & CueSet & 0.68 & 0.63 & 0.71 & 0.70 & 0.67 \\
{\em VAL} & CertSet & 0.76 & 0.79 & 0.96 & 0.74 & 0.67 \\
\hline
\end{tabular}
\end{center}
\caption{Average results after 10-fold cross validation on held-out data for the two tasks: tweet resolution ({\em RES}) and resolution value ({\em VAL}) prediction and the two feature sets {\em CueSet} (lexical cue ratios) and {\em CertSet} (predicted certainty values).}
\label{tab:res_pred}
\end{table}

\section{Discussion and conclusion}
\label{sec:conc}

\paragraph{Relation between lexical cues and certainty}

As described in Section \ref{sec:facPred}, we established a link between lexical cue ratios of different certainty levels and the certainty associated to tweets by means of regression analysis.
The zero-inflation problem 
 as well as the reported low correlations between each cue ratio and the certainty values indicate that factuality values cannot fully be expressed by cue-type ratios in isolation but require a more complex model. Applying GLMs to bundle and therefore strengthen these weak relations was a first step in this direction.
Certainty is a discourse-level phenomenon that 
lexical means can represent to some extent but not entirely.
In future work we are going to address the representation of certainty phenomena related to higher linguistic levels.

\paragraph{Impact of rumor resolution on cue ratios and certainty}

As pointed out in Section \ref{sec:facDriftQuant}, for several examined cue ratio and certainty variables significant differences were observed with respect to claim resolution and resolution value. 
Resolving tweets show higher knowledge ({\em KCR}) and report cue ratios ({\em RCR}) as well as related discontinuities. Positive resets mark an abrupt increase in knowledge and report cues at the time point of resolution. 
This is even more pronounced when combining both cue ratio related features
to a common one, the factuality cue ratio 
({\em FCR}= {\em KCR} + {\em RCR}). 
Claim resolution has a major impact on the distribution of lexical cues associated with a high certainty level, as after resolution their amount increases.
Belief cue ratios, on the contrary, have not proven to be of relevance for distinguishing between resolving and non-resolving tweets.
Doubt cue and certainty variables are additionally highly dependent on the claim resolution value.
Doubt cues occur more often in falsified claims and show a higher increase after claim resolution when compared to verified claims ({\em DCR\_Delta, DCR\_Reset, DCR\_RMS\_f}). Certainty cues behave exactly the opposite way: they are generally more frequent in verified claims ({\em CRT}) and at the resolution point they show an upward shift in verified but a downward shift in falsified claims ({\em CRT\_Reset}). These interpretations are further supported by the {\em RES-VAL} interactions for the doubt cue and certainty variables, i.e. the amount and the direction of discontinuity at the resolution point of doubt cues and certainty values depends on whether the claim is verified or falsified.

\paragraph{Prediction of resolution and its value}

Table \ref{tab:res_pred} shows that the prediction tasks benefit from the feature sets in different ways. For task {\em RES}, the lexical cue set {\em CueSet} turned out to be more appropriate, while for {\em VAL} the certainty values {\em CertSet}. The reason might be that for {\em VAL} only resolving tweets and thus a lower amount of data is available suggesting the 
utility of shorter vectors containing derived instead of raw features. Importantly for sparse data scenarios, the generation of such an intermediate level of cue-integrating features, in 
our case the predicted certainty, turns out to be beneficial.

\newpage

\bibliographystyle{acl}
\bibliography{piro_newbib}

\end{document}